\documentclass[letterpaper]{article} % DO NOT CHANGE THIS
\usepackage{aaai2026}  % DO NOT CHANGE THIS
\usepackage{times}  % DO NOT CHANGE THIS
\usepackage{helvet}  % DO NOT CHANGE THIS
\usepackage{courier}  % DO NOT CHANGE THIS
\usepackage[hyphens]{url}  % DO NOT CHANGE THIS
\usepackage{graphicx} % DO NOT CHANGE THIS
\usepackage{natbib}  % DO NOT CHANGE THIS AND DO NOT ADD ANY OPTIONS TO IT
\usepackage{caption} % DO NOT CHANGE THIS AND DO NOT ADD ANY OPTIONS TO IT
\usepackage{algorithm}
\usepackage{algpseudocode}
\usepackage{booktabs}   % For professional table lines
\usepackage{tabularx}   % For tables with auto-adjusting 
\usepackage{amsfonts}
\usepackage{amssymb}
\usepackage{graphicx}

\usepackage{array}
\usepackage{makecell}   % For multi-line headers (\thead)
\usepackage{multirow}   
\usepackage{siunitx}

\newcolumntype{C}{>{\centering\arraybackslash}X}
\usepackage{newfloat}
\usepackage{listings}
\DeclareCaptionStyle{ruled}{labelfont=normalfont,labelsep=colon,strut=off} % DO NOT CHANGE THIS
\floatstyle{ruled}
\newfloat{listing}{tb}{lst}{}
\floatname{listing}{Listing}
\usepackage{booktabs}
\usepackage{tabularx}
\usepackage{amsmath}
\usepackage{amsthm}
\usepackage{subcaption}
\usepackage{bbding}
\newtheorem{theorem}{Theorem}% 
\newtheorem{remark}{Remark}% 
\newtheorem{lemma}{Lemma}%

\nocopyright 
\title{Hierarchical Schedule Optimization for Fast and Robust Diffusion Model Sampling}
\author {
    Aihua Zhu\textsuperscript{\rm 1}\equalcontrib,
    Rui Su\textsuperscript{\rm 1}\equalcontrib,
    Qinglin Zhao\textsuperscript{\rm 1}\thanks{Corresponding author.},
    Li Feng\textsuperscript{\rm 1},
    Meng Shen\textsuperscript{\rm 2},
    Shibo He\textsuperscript{\rm 3}
}
\affiliations {
    \textsuperscript{\rm 1}School of Computer Science and Engineering, Macau University of Science and Technology, Macau 999078, China\\
    \textsuperscript{\rm 2}Beijing Institute of Technology\\
    \textsuperscript{\rm 3}Zhejiang University\\
    aihuaz@student.must.edu.mo, ruis10@student.must.edu.mo,
    qlzhao@must.edu.mo, lfeng@must.edu.mo,
    shenmeng@bit.edu.cn, s18he@zju.edu.cn
}

\usepackage{bibentry}
\begin{document}

\maketitle

\begin{abstract}
Diffusion probabilistic models have set a new standard for generative fidelity but are hindered by a slow iterative sampling process. A powerful training-free strategy to accelerate this process is \textit{Schedule Optimization}, which aims to find an optimal distribution of timesteps for a fixed and small Number of Function Evaluations (NFE) to maximize sample quality. To this end, a successful schedule optimization method must adhere to four core principles: effectiveness, adaptivity, practical robustness, and computational efficiency. However, existing paradigms struggle to satisfy these principles simultaneously, motivating the need for a more advanced solution. To overcome these limitations, we propose the Hierarchical-Schedule-Optimizer (HSO), a novel and efficient bi-level optimization framework. HSO reframes the search for a globally optimal schedule into a more tractable problem by iteratively alternating between two synergistic levels: an upper-level global search for an optimal initialization strategy and a lower-level local optimization for schedule refinement. This process is guided by two key innovations: the Midpoint Error Proxy (MEP), a solver-agnostic and numerically stable objective for effective local optimization, and the Spacing-Penalized Fitness (SPF) function, which ensures practical robustness by penalizing pathologically close timesteps. Extensive experiments show that HSO sets a new state-of-the-art for training-free sampling in the extremely low-NFE regime. For instance, with an NFE of just 5, HSO achieves a remarkable FID of 11.94 on LAION-Aesthetics with Stable Diffusion v2.1. Crucially, this level of performance is attained not through costly retraining, but with a one-time optimization cost of \textit{less than 8 seconds}, presenting a highly practical and efficient paradigm for diffusion model acceleration.

\end{abstract}

\begin{links}
    \link{Code}{https://github.com/chappy0/HSO.git}

\end{links}
% Uncomment the following to link to your code, datasets, an extended version or similar.
% You must keep this block between (not within) the abstract and the main body of the paper.
% \begin{links}
%     \link{Code}{https://aaai.org/example/code}
%     \link{Datasets}{https://aaai.org/example/datasets}
%     \link{Extended version}{https://aaai.org/example/extended-version}
% \end{links}

\section{Introduction}
\label{sec:introduction}

\begin{table}[t]
\centering
\small
\renewcommand{\arraystretch}{1.5}
\begin{tabular}{l|ccc|c}
\hline
\textbf{Principle} & \makecell{\textbf{Rule-}\\\textbf{based}} & \makecell{\textbf{Perceptual}\\\textbf{Opt.}} & \makecell{\textbf{Principled}\\\textbf{Opt.}} & \makecell{\textbf{HSO}\\\textbf{(Ours)}} \\
\hline
Adaptivity & \XSolidBrush & \Checkmark & \Checkmark & \Checkmark \\
Effectiveness & \XSolidBrush & \Checkmark & \XSolidBrush & \Checkmark \\
\makecell{Practical\\Robustness} & \Checkmark & \XSolidBrush & \XSolidBrush & \Checkmark \\
\makecell{Comp.\\Efficiency} & \Checkmark & \XSolidBrush & \Checkmark & \Checkmark \\
\hline
\end{tabular}
\caption{A comparison of Schedule Optimization paradigms on four key principles. HSO uniquely satisfies all four.}
\label{tab:paradigm_comparison}
\end{table}

Diffusion probabilistic models \citep{ho2020denoising, song2021scorebased} have established a new standard for high-quality image generation. However, their practical utility is hindered by a computationally intensive iterative sampling process. At the heart of this process from pure noise to a clean image, a numerical solver repeatedly invokes the model's neural network function over a sequence of discrete timesteps, known as the \textit{schedule}. The total number of these invocations termed the Number of Function Evaluations (NFE), often reaches the hundreds or thousands, creating a significant bottleneck for real-time applications. Consequently, this makes the reduction of NFE while maintaining high sample fidelity a primary research goal.

A powerful strategy to address this goal is \textit{Schedule Optimization}. This approach focuses on determining the optimal distribution of a limited set of sampling timesteps across the full generation process from noise to image, which in turn defines the sampling schedule. A key advantage is its \textbf{training-free} nature, which refines the sampling process to accelerate diffusion models without costly retraining or fine-tuning. The ultimate objective of this task is to discover a schedule that yields the highest possible sample fidelity with a minimal search cost. To systematically design an approach capable of meeting this objective, we argue that it must adhere to four fundamental principles:

%A key advantage of this direction is that it is inherently \textbf{training-free}, which, in the context of diffusion models, means refining the sampling process without any costly retraining or fine-tuning of the model itself, making it a flexible and efficient path to acceleration. 

\begin{itemize}
    \item \textbf{Adaptivity}: The process must derive a schedule adapted to the intrinsic characteristics of models and given NFE budgets, rather than relying on a one-size-fits-all rule independent to the models and the NFE.
    
    \item \textbf{Effectiveness}: The underlying logic of the method must be capable of producing a schedule that results in high-fidelity samples.

    \item \textbf{Practical Robustness}: The approach must incorporate explicit constraints to ensure robustness against the deviation between its theoretical objective and actual fidelity.

    \item \textbf{Computational Efficiency}: The total cost of the search process must be practically viable and acceptable.
    
\end{itemize}

However, as summarized in Table \ref{tab:paradigm_comparison}, current paradigms for schedule optimization struggle to satisfy these four principles simultaneously, each presenting its own trade-offs. \textbf{Rule-based methods}, while computationally efficient and often robust, inherently lack Adaptivity as their fixed rules cannot adjust to diverse models and scenarios. On the other hand, optimization-based approaches are adaptive but struggle with the other principles. \textbf{Perceptual Optimization}, which directly optimizes for image perceptual quality metrics, sacrifices Computational Efficiency due to the high cost of repeated image generation during the search. \textbf{Principled Optimization}, which \textit{seeks to minimize a theoretically-derived objective} instead, are undermined by a non-convex landscape that traps local searches in suboptimal minima, crippling their Effectiveness and often neglecting Practical Robustness.

To resolve these shortcomings and holistically satisfy all aforementioned principles, we introduce the \textbf{Hierarchical-Schedule-Optimizer (HSO)}, a novel and efficient bi-level optimization framework designed to navigate the non-convex search landscape. HSO iteratively alternates between an upper-level global search within a low-dimensional space for an optimal initialization strategy and a lower-level local optimization for schedule refinement from that superior initial point. This process is guided by two key innovations: the \textbf{Midpoint Error Proxy (MEP)}, a solver-agnostic and numerically stable objective for effective local optimization, and the \textbf{Spacing-Penalized Fitness (SPF)} function, which ensures practical robustness by penalizing pathologically close timesteps.

Our main contributions can be summarized as follows: 
\begin{itemize}
   
    \item We propose HSO, a novel bi-level optimization framework that significantly advances the Principled Optimization paradigm, with a search process that adheres to the principles of Adaptivity, Effectiveness, Practical Robustness, and Computational Efficiency.

    \item Within the HSO framework, we present two key technical innovations: the MEP guiding the lower-level local optimization, and the SPF function steering the upper-level global search.

    \item We demonstrate through extensive experiments that our approach sets a new state-of-the-art for training-free sampling in the extremely low-NFE regime. For instance, at an NFE of 5, using the Stable Diffusion v2.1 model on the LAION-Aesthetics 6.5+ benchmark, HSO achieves an FID\citep{heusel2017gans} score as low as 11.94, all with a one-time optimization cost of \textbf{less than 8 seconds}, presenting a highly practical paradigm for diffusion model acceleration.
    
\end{itemize}

\section{Related Work}
\label{sec:related_work}
Accelerating diffusion model sampling is a critical area of research. To precisely position our contributions, we first overview the broader acceleration landscape before focusing on prior work in Schedule Optimization.

\subsection{Approaches to Sampling Acceleration}
Methods for accelerating diffusion model sampling can be broadly divided by whether they require retraining the diffusion model itself.

\textbf{Training-based acceleration} creats or fine-tunes a diffusion model for few-step sampling. Prominent techniques include knowledge distillation \citep{geng2023one,salimans2022progressive,sauer2024adversarial,luo2023diff,yin2024one,zhou2024score,frans2025one,lin2024sdxl,xu2024ufogen,watson2021learning1,watson2021learning2}, consistency models \citep{song2023consistency,geng2024consistency,lu2025simplifying,luo2023latent,song2023improved,yang2024consistency}, and the advanced work Mean Flows \citep{geng2025mean}, which can achieve extremely low NFE, even single-step generation. Their primary drawback, however, is the substantial computational cost required for this training or fine-tuning process, which limits their practical applicability to new or existing models. \textbf{Training-free acceleration}, aims to speed up any pre-trained model without modifying its weights. This area encompasses: 1) Solver Design, which focuses on creating custom ODE/SDE solvers or using advanced test-time guidance\citep{zhao2023unipc,lu2022dpm, lu2025dpm,frankel2025s4s,gonzalez2023seeds,wang2025pfdiff,xia2023towards,kim2025test}; 2) Model-level, which optimizes the neural network's architecture through techniques like quantization or caching\citep{shang2023post,liu2025cachequant}; 3) Schedule Optimization, the focus of our work.

\subsection{Schedule Optimization Strategies}
\label{sec:so}
The task of schedule optimization is to select an optimal subsequence of timesteps. Existing strategies are typically either rule-based or optimization-based.

\textbf{Rule-based schedules} provide a fixed, pre-defined formula for generating timesteps, shuch as the power-law distribution used in the EDM schedule \citep{karras2022elucidating} and the uniform log-SNR schedule defaulted by DPM-Solver \citep{lu2022dpm}. \textbf{Optimization-based schedules} formulate the task as a search problem, aiming to find a schedule that optimizes a specific objective function. This category can be further divided into two main branches: \textbf{perceptual optimization} methods directly optimize for image quality metrics such as FID or KID \citep{li2023autodiffusion,liu2023oms,wang2023learning}; \textbf{principled optimization} method like DM-NonUni \citep{xue2024accelerating} and Score-Optimal Diffusion Schedules \citep
{williams2024score}, which is most similar to our work, aim to minimize a theoretically-derived objective.

\section{Preliminaries: Foundations of Principled Schedule Optimization}
\label{sec:preliminaries}
This section establishes the technical foundations for Principled Schedule Optimization by defining the schedule optimization problem within the probability flow ODE framework, introducing a parameterized method for constructing an initial schedule, and outlining the principle of a theoretical optimization objective.

\subsection{The Schedule Optimization Problem}
\label{sec:probelm_setup}
A diffusion model's forward process transforms a clean data sample $\mathbf{x}_0$ into pure noise over a continuous time $t\in[0,T]$. The distribution of the noisy sample $\mathbf{x}_t$ conditioned on $\mathbf{x}_0$ is a Gaussian: $q(\mathbf{x}_t|\mathbf{x}_0)=\mathcal{N}(\alpha_t\mathbf{x}_0, \sigma_t^2\mathbf{I})$, where $\alpha_t$ and $\sigma_t$ are the signal and noise scales, respectively.

The reverse generation process can be described by a probability flow Ordinary Differential Equation (ODE) for deterministic sampling\citep{song2021scorebased}. For a model with a data prediction network $\mathbf{x_\theta}(\mathbf{x}_t, t)$ to approximate the original clean data $\mathbf{x}_0$, this ODE takes the specific form \cite{xue2024accelerating}:
\begin{equation}
\label{eq:prob_flow_ode}
\frac{\mathrm{d} \mathbf{x}_t}{\mathrm{d}t} = \frac{\mathrm{d}\log\sigma_t}{\mathrm{d}t}\mathbf{x}_t + \frac{\alpha_t\mathrm{d}\log\frac{\alpha_t}{\sigma_t}}{\mathrm{d}t}\mathbf{x_\theta}(\mathbf{x}_t, t)
\end{equation}

A numerical ODE solver approximates the solution to this equation by evaluating $\mathbf{x_\theta}$ at a discrete sequence of timesteps $\tau = \{t_0, t_1, \cdots, t_N\}$, where $t_0=T$ is the start time and $t_N=\epsilon$ is the end time for some small $\epsilon>0$. For improved numerical stability, this process is often performed in the log-signal-to-noise ratio (log-SNR) space, defined by $\lambda_t = \log(\alpha_t / \sigma_t)$. The core problem of schedule optimization is therefore to find the optimal discrete set of log-SNR values, $\Lambda = \{\lambda_0, \lambda_1, \cdots, \lambda_N\}$, that minimizes the final generation error for a fixed NFE budget $N$.

\subsection{Parameterizing the Optimization Initial Point}
\label{sec:prelim_edm}
Optimization-based method requires an initial schedule $\Lambda_\text{init}$, or a initial point, from which to begin its search. A structured way to construct this schedule is by generating it from a low-dimensional hyperparameter vector, denoted by $\boldsymbol{\psi}$. The highly effective EDM schedule proposed by \citeauthor{karras2022elucidating}(\citeyear{karras2022elucidating}) provides an excellent formula for this purpose. Their method defines the schedule based on a noise level parameter, $\tilde{\sigma}$, which relates to the total noise and signal scales via $\sigma_t = \alpha_t\tilde{\sigma}_t$. The schedule for these discrete noise level parameters $\{\tilde{\sigma}_i\}_{i=0}^N$ is constructed by the formula($\tilde{\sigma}_N = 0$):

\begin{equation}
\label{eq:karras_schedule}
\tilde{\sigma}_i = \left( \tilde{\sigma}_{\text{min}}^{1/\rho} + \frac{i}{N-1} \left( \tilde{\sigma}_{\text{max}}^{1/\rho} - \tilde{\sigma}_{\text{min}}^{1/\rho} \right) \right)^{\rho}
\end{equation}

This entire construction is governed by the compact hyperparameter vector $\boldsymbol{\psi}=(\rho, \tilde{\sigma}_\text{min},\tilde{\sigma}_\text{max})\in \mathbb{R}^3$, which defines the schedule's distribution and boundaries. By converting these $\tilde{\sigma}_i$ values to the log-SNR space via $\lambda_i = -\log(\tilde{\sigma}_i)$, we can generate a complete log-SNR schedule, denoted as $\Lambda(\boldsymbol{\psi})$. This parameterized function allows a complete initial schedule, $\Lambda_\text{init}=\Lambda(\boldsymbol{\psi})$, to be generated from a compact set of inputs. The low-dimensional vector $\boldsymbol{\psi}$ thus represents the ``initialization strategy" that HSO's upper level is designed to optimize.

\subsection{Principle of a Theoretical Optimization Objective}
\label{sec:prelim_j_theo}
Principled schedule optimization methods require a theoretical objective function, $\mathcal{J}_\text{theo}(\Lambda)$, to guide the search for an optimal schedule. The core idea, pioneered by DM-NonUni, is to derive this objective from the global generation error $||\tilde{x}_{\epsilon}-x_0||$, where $\tilde{x}_{\epsilon}$ is the numerical approximation of the true solution $x_\epsilon$. This error originates from the integral solution of the ODE in equation \ref{eq:prob_flow_ode}:
\begin{equation}
\label{eq:ODE_solution}
    x_\epsilon = \frac{\sigma_\epsilon}{\sigma_T}x_T+\sigma_\epsilon\sum_{i=0}^{N-1}\int_{\lambda_i}^{\lambda_{i+1}}e^\lambda f(\lambda)d\lambda
\end{equation}
where $f(\lambda) = \mathbf{x}_\theta(\mathbf{x}_{t(\lambda)},t(\lambda))$.
A tractable objective can be formulated by first approximating the integral term. The specific form of this objective depends on the chosen approximation method for $f(\lambda)$ within the integral. This principle provides the theoretical foundation for the novel objective function we introduce in our Section \ref{sec:mep}.

\section{Methodology: The HSO}
\label{sec:methodology}
This section details the Hierarchical-Schedule-Optimizer (HSO), our framework designed to systematically satisfy the four foundational principles of schedule optimization. We will first present its general bi-level architecture, then detail its two core components, the Midpoint Error Proxy (MEP) and the Spacing-Penalized Fitness (SPF), before concluding with the complete algorithm.

\subsection{The HSO Framework: A Bi-Level Solution to a Non-Convex Problem}
\label{sec:mso}
The goal of principled schedule optimization is to find a schedule $\Lambda$ that minimizes a theoretical objective function $\mathcal{J}_\text{theo}(\Lambda)$ derived from the global generation error in Section \ref{sec:prelim_j_theo}. However, the optimization landscape $\mathcal{J}_\text{theo}(\Lambda)$ is highly non-convex, making local optimization methods highly sensitive to their initial point $\Lambda_\text{init}$ and prone to getting trapped in suboptimal minima \citep{jain2017non}.

%, an observation supported by empirical evidence in Section \ref{sec:non-convex-and-nfe-adaptive}). 

The direct consequence of this non-convex landscape is that the success of local optimization becomes critically dependent on finding a superior initial point. While a global search is the most robust method for locating such a point, searching for a schedule $\Lambda_\text{init}\in \mathbb{R}^N$ directly in the $N$-dimensional schedule space is often computationally intractable due to the curse of dimensionality, especially as the NFE $N$ increases. To overcome this tractability barrier, HSO reframes the problem from a direct, high-dimensional search over $\Lambda$ to a more tractable, global search for a single point $\boldsymbol{\psi}$ that generates the initial schedule within the low-dimensional hyperparameter space $\mathbb{R}^3$. This reframing naturally gives rise to the HSO bi-level framework, which decomposes the difficult optimization for a global optimal schedule into two manageable sub-problems: a global search for an optimal initialization strategy and a local optimization of the resulting schedule, which are solved through an iterative, alternating process between the two levels of our framework:

\begin{itemize}
    \item \textbf{Lower-Level Local Optimization}: This level performs a local search for an optimal schedule $\Lambda_\text{opt}$, starting from an initial point $\Lambda_\text{init}$ which is generated using the hyperparameter vector $\boldsymbol{\psi}$ from the upper level. This search is guided by a flexible theoretical objective function $\mathcal{J}_\text{lower}$ and executed using a standard constrained optimization algorithm, such as the Trust-Region Constrained Optimization method \citep{conn2000trust}. Formally, its task is to solve:
    \begin{equation}
        \Lambda_\text{opt}(\boldsymbol{\psi}) = \arg\min_\Lambda \mathcal{J}_\text{lower}(\Lambda|\Lambda_\text{init}(\boldsymbol{\psi}))
    \end{equation}

    \item \textbf{The Upper-level Global Search}: This level performs a global search for the optimal hyperparameter vector $\boldsymbol{\psi}^*$ that defines the best initial point. It employs a population-based evolutionary algorithm, such as Differential Evolution \citep{storn1997differential}, to effectively navigate the non-convex landscape. This search is guided by a flexible, practicality-aware fitness function $\mathcal{F}_\text{upper}$, which evaluates the quality of the final schedule $\Lambda_\text{opt}(\boldsymbol{\psi})$ produced by the lower level for a given $\boldsymbol{\psi}$. The task can be formally expressed as:
    \begin{equation}
        \boldsymbol{\psi}^* = \arg\min_{\boldsymbol{\psi}}\mathcal{F}_\text{upper}(\Lambda_\text{opt}(\boldsymbol{\psi}))
    \end{equation}

\end{itemize}
This bi-level optimization is realized through an iterative, alternating process. In each cycle, the upper level proposes a population of candidate strategies $\{\boldsymbol{\psi}_j\}$. For each candidate, the lower level executes a local optimization to yield a corresponding schedule $\{\Lambda_\text{opt,j}\}$. The quality of these schedules is then evaluated and fed back to the upper level to guide the evolution of its population for the next cycle. This iterative process of global search and local optimization continues until a termination condition is met.

The performance of the HSO framework is ultimately driven by the specific designs of its two guiding functions, the lower-level objective $\mathcal{J}_\text{lower}$ and the upper-level fitness function $\mathcal{F}_\text{upper}$, which we detail in the following subsections.

\subsection{Objective for Local Optimization: The Midpoint Error Proxy (MEP)}
\label{sec:mep}
The design of the lower level's theoretical objective $\mathcal{J}_\text{lower}$, critically influences the Effectiveness, Adaptivity and Computational Efficiency of the HSO framework. While DM-NonUni derives objectives tightly coupled to the mechanics of specific high-order solvers like UniPC \citep{zhao2023unipc}, this approach can limit generality and numerical stability, as the underlying extrapolation-based polynomial integral approximation used in UniPC sampling can produce large estimation errors and lead to instability\citep{burden1993numerical}. As a core theoretical contribution of this work, we diverge from this solver-specific path and propose a more universal and numerically reliable objective, the Midpoint Error Proxy (MEP) $\mathcal{J}_\text{MEP}$, to serve as the lower level's objective $\mathcal{J}_\text{lower}$.

Following the principle from Section \ref{sec:prelim_j_theo}, our goal is to derive a tractable objective by approximating the integral term in the global error formula shown in Eq.(\ref{eq:ODE_solution}). Our approach is use an interpolation-based approximation inspired by the classic midpoint rule, based on the principle of isolating and exactly integrating the analytically tractable exponential term $e^\lambda$, while only applying approximation to the neural network term $f(\lambda)$ which we formalize in Lemma \ref{lemma:approx}.

\begin{lemma}[Hybrid Midpoint Approximation]
The integral term can be approximated as: 
\begin{equation}
\label{eq:integral_approx}
    \int_{\lambda_i}^{\lambda_{i+1}}e^\lambda f(\lambda)d\lambda \approx f(\lambda_{i+\frac{1}{2}})(e^{\lambda_{i+1}}-e^{\lambda_i})
\end{equation}
where $\lambda_{i+\frac{1}{2}}\overset{\triangle}{=} \frac{\lambda_i+\lambda_{i+1}}{2}$. The local truncation error of this approximation is $O(h^3)$, the same order as the standard midpoint rule, but it is numerically more stable for this specific integral structure.
\label{lemma:approx}
\end{lemma}
A detailed derivation and proof are provided in the Appendix A. Building on this lemma, we can now establish the global error bound. To formalize this bound, we adopt the theoretical setup from DM-NonUni, assuming the model's prediction is $f(\lambda)=x_0+\xi_{t(\lambda)}$, where the error norm is bounded with high probability by $||\xi_{t(\lambda)}||\leq\tilde{\eta}\tilde{\epsilon}_{t(\lambda)}$ for some constant $\tilde{\eta}$ and error proxy $\tilde{\epsilon}_{t(\lambda)}=\frac{\sigma_t^p}{\alpha_t}$ for a non-negative integer $p$.

\begin{theorem}[Global Error Bound for Hybrid Midpoint Approximation]
\label{theo:error}
Under the aforementioned assumptions, let the approximate solution $\tilde{x}_{\epsilon,MEP}$ be derived using the approximation from Lemma  \ref{lemma:approx}. The global generation error is then bounded with high probability by:
    \begin{equation}
        ||\tilde{x}_{\epsilon,MEP}-x_0||\leq C + \sigma_\epsilon\tilde{\eta}\sum_{i=0}^{N-1}\tilde{\epsilon}_{t(\lambda_{i+\frac{1}{2}})}(e^{\lambda_{i+1}}-e^{\lambda_i})
    \end{equation}
    where $C$ represents schedule-independent error terms.
\end{theorem}

A proof is provided in the Appendix B. Based on Theorem \ref{theo:error}, minimizing the error bound is achieved by minimizing the schedule-dependent summation. This directly yields our MEP objective:
\begin{equation}
    \mathcal{J}_\text{MEP}(\Lambda) = \sum_{i=0}^{N-1}\tilde{\epsilon}_{t(\lambda_{i+\frac{1}{2}})}(e^{\lambda_{i+1}}-e^{\lambda_i})
\end{equation}

\begin{remark}[Adherence to Core Principles]
The MEP objective is deliberately designed to embody our core principles. Its high-order accuracy and numerical reliability ensure \textbf{Effectiveness}. It ensures \textbf{Adaptivity} because the error proxy $\tilde{\epsilon}_{t(\lambda)}$ incorporates the model's unique noise schedule parameters and the optimization is performed for a specific NFE. Finally, its clean analytical form is computable in linear time $O(N)$, ensures \textbf{Computational Efficiency}.
\end{remark}

\subsection{Fitness Function For Global Search: The Spacing-Penalized Fitness (SPF)}
\label{sec:fitness}

The design of the upper level's fitness function $\mathcal{F}_\text{upper}$ is crucial for ensuring the Practical Robustness of the discovered schedules. Our empirical analysis reveals that unconstrained optimization can produce schedules with pathologically close timesteps. While theoretically optimal, these schedules are practically ineffective, as the close steps provide negligible progress and waste the NFE budget.

To address this, we design the Spacing-Penalized Fitness (SPF) $\mathcal{F}_\text{SPF}$ to serve as $\mathcal{F}_\text{upper}$. It augments the theoretical error from our MEP objective with a dynamic spacing penalty, $L_\text{penalty}$. The combined fitness function is:

\begin{equation}
\mathcal{F}_\text{SPF}(\psi|N) = \mathcal{J}_\text{MEP}(\Lambda_\text{opt}(\boldsymbol{\psi})) + \gamma L_\text{penalty}(\Lambda_\text{opt}(\boldsymbol{\psi})|N)
\end{equation}
where the penalty term heavily penalizes schedules when timesteps cluster closer than a minimum distance $d_\text{min}(N)$ that adapts based on the NFE budget:
\begin{equation}
    L_\text{penalty} = \sum_{i=0}^{N-1}\max(0, d_\text{min}(N)-|t(\lambda_{i+1})-t(\lambda_i)|)^2
\end{equation}
The value of $d_\text{min}(N)$ adapts to the NFE budget based on a simple linear heuristic, decreasing from 0.15 at NFE $N=4$ down to 0.01 at $N=20$. This strategy enforces a wider step spacing to ensure robustness at very low NFEs, while permitting finer-grained steps for higher precision when a larger evaluation budget is available.

\subsection{HSO Implementation and Algorithm}
\label{sec:agr}
We summarize the complete HSO algorithm, built on our core components MEP and SPF, in Algorithm \ref{alg:HSO}. The synergy between its efficient, low-dimensional global search and robust, MEP-guided local optimization objective regularized by the SPF's spacing penalty, ensures the discovery of schedules that are simultaneously Effective, Adaptive, Practically Robust, and Computationally Efficient.

\begin{algorithm}[tb]
\caption{The Complete HSO Optimization Process}
\label{alg:HSO}
\textbf{Input:} Target NFE $N$, Search bounds for strategy vector $\boldsymbol{\psi}$\\
\textbf{Output:} Optimized schedule $\Lambda^*$
\begin{algorithmic}[1]

\State Initialize a set of candidate hyperparameter vectors $P=\{\boldsymbol{\psi}_1, \dots, \boldsymbol{\psi}_K\}$ \Comment{Init each $\boldsymbol{\psi}_k\in \mathbb{R}^3$}
\State Initialize best fitness $f^* \leftarrow \infty$, best schedule $\Lambda^* \leftarrow \text{null}$
\While{\texttt{TermCond} is not met} \Comment{// Global Search}
    \For{each candidate $\boldsymbol{\psi}_k$ in $P$}\Comment{// Local Opt.}
        \State $\Lambda_\text{init} \leftarrow \text{GenerateInitialPoint}(\boldsymbol{\psi_k})$
        \State $\Lambda_\text{opt,k} \leftarrow \text{Minimize}(\mathcal{J}_{MEP}, \Lambda_\text{init})$
        \Statex \Comment{// Minimize $\mathcal{J}_{MEP}$ with $\Lambda_\text{init}$ as start}
        \State $f_k \leftarrow \mathcal{F}_\text{SPF}(\Lambda_\text{opt,k})$ \Comment{// Evaluate fitness of $\Lambda_\text{opt,k}$}
        \If{$f_j < f^*$}
            \State $f^* \leftarrow f_k$
            \State $\Lambda^* \leftarrow \Lambda_\text{opt,k}$
        \EndIf
    \EndFor
    \State $P\leftarrow \text{EvolvePopulation}(P,\{f_1,\dots,f_K\})$
\EndWhile
\end{algorithmic}
\end{algorithm}

\section{Experiments}
\label{sec:experiments}
This section experimentally validates our Hierarchical Schedule Optimizer (HSO). Following the setup, we demonstrate its Effectiveness, Adaptivity, Practical Robustness, and Computational Efficiency. Finally, an ablation study verifies the individual and synergistic contributions of HSO’s core components.

\subsection{Experimental Setup}

For reproducibility, we evaluate HSO on the Stable Diffusion v2.1-base model, a widely-adopted open-source model, using two prominent ODE solvers, UniPC \citep{zhao2023unipc} and DDIM \citep{song2020denoising}, assessing its generalization capabilities across three standard, large-scale benchmarks: LAION-Aesthetics 6.5+ (comprising approximately 30,000 text-image pairs)\cite{schuhmann2022laion}, the MS-COCO 2017 validation set (with 30,000 captions)\citep{lin2014microsoft}, and the ImageNet 512x512 validation set (with 50,000 captions)\citep{deng2009imagenet}. We measure image quality using the Fréchet Inception Distance (FID) \citep{heusel2017gans}, a standard metric for perceptual fidelity. The main parameters for HSO, espetially the search boundaries for the strategy vector $\boldsymbol{\psi}$ and our rationale for this, are detailed in the Appendix C. We benchmark HSO against SOTA acceleration methods, especially the principled optimization DM-NonUni\citep{xue2024accelerating} focusing on the demanding low-NFE ($N < 5$) regime.

\subsection{Validation of the Effectiveness}

\textbf{Quantitative analysis.}
To validate its effectiveness, we use FID\citep{heusel2017gans} to quantitatively measure sample fidelity, where a lower score indicates higher quality. As shown in Table \ref{tab:HSO_expected_performance}, HSO consistently and significantly outperforms the baseline DM-NonUni across all configurations. This advantage is most pronounced in the extremely low-NFE regime. For instance, on LAION-Aesthetics with the DDIM solver, as NFE drops from 5 to 4, the baseline FID collapses from 35.38 to 68.92, whereas HSO's degrades more gracefully from 17.17 to 24.77. While HSO shows clear improvements on UniPC, the performance gain is substantially larger on the DDIM solver (e.g., on LAION, FID drops from 68.92 to 24.77 at $NFE=4$). This is because the baseline's objective was tailored for UniPC, whereas our solver-agnostic MEP objective generalizes more effectively, thereby validating the universality of our approach. This robust performance across different solvers, especially in the demanding low-NFE regime, confirms HSO's ability to find highly effective schedules.

\begin{table}[htbp]
\centering
\setlength{\tabcolsep}{4.8pt}
\renewcommand{\arraystretch}{1.12}
\resizebox{\columnwidth}{!}{
\begin{tabular}{@{} p{1.7cm} p{0.9cm} c *2{>{\centering\arraybackslash}p{1.9cm}} @{}}
\toprule
\multicolumn{1}{c}{\textbf{Dataset}} &
\multicolumn{1}{c}{\textbf{Solver}} &
\multicolumn{1}{c}{\textbf{NFE}} &
\multicolumn{1}{c}{\textbf{FID$\downarrow$ (CVPR)}} &
\multicolumn{1}{c}{\textbf{FID$\downarrow$ HSO (Ours)}} \\
\midrule
\multirow{4}{=}{\centering LAION-Aesthetics\\6.5+}
 & \multirow{2}{*}{\centering UniPC} & 4 & 18.96 & \textbf{15.71} \\
 & & 5 & 13.91 & \textbf{11.94} \\
\cline{2-5}
 & \multirow{2}{*}{\centering DDIM} & 4 & 68.92 & 24.77 \\
 & & 5 & 35.38 & 17.17 \\
\midrule
\multirow{4}{=}{\centering MS-COCO}
 & \multirow{2}{*}{\centering UniPC} & 4 & 27.50 & 23.26 \\
 & & 5 & 23.10 & 19.59 \\
\cline{2-5}
 & \multirow{2}{*}{\centering DDIM} & 4 & 60.06 & 29.55 \\
 & & 5 & 30.12 & 23.15 \\
\midrule
\multirow{4}{=}{\centering ImageNet\\(512$\times$512)}
 & \multirow{2}{*}{\centering UniPC} & 4 & 20.75 & 17.20 \\
 & & 5 & 17.63 & 15.90 \\
\cline{2-5}
 & \multirow{2}{*}{\centering DDIM} & 4 & 41.51 & 19.78 \\
 & & 5 & 22.89 & 16.02 \\
\bottomrule
\end{tabular}}
\caption{Performance comparison (FID$\downarrow$) between HSO and DM-NonUni. The ``FID (CVPR)" column refers to the results with the DM-NonUni method \cite{xue2024accelerating}.}
\label{tab:HSO_expected_performance}
\end{table}

\begin{figure}[htbp]
\centering

\label{fig:qualitative_comparison}

\begin{tabular*}{\linewidth}{@{\extracolsep{\fill}}ccc@{}}
    \small
    \parbox[t]{0.31\linewidth}{\centering \textbf{LAION}:\\A cat wearing an astronaut helmet.\\NFE=4} &
        \parbox[t]{0.31\linewidth}{\centering \textbf{COCO}:\\A vintage red sports car.\\NFE=5} &
        \parbox[t]{0.31\linewidth}{\centering \textbf{ImageNet}:\\Golden\\ retriever\\NFE=4} \\
\end{tabular*}
    
\begin{subfigure}{0.3\linewidth}
    \includegraphics[width=\linewidth]{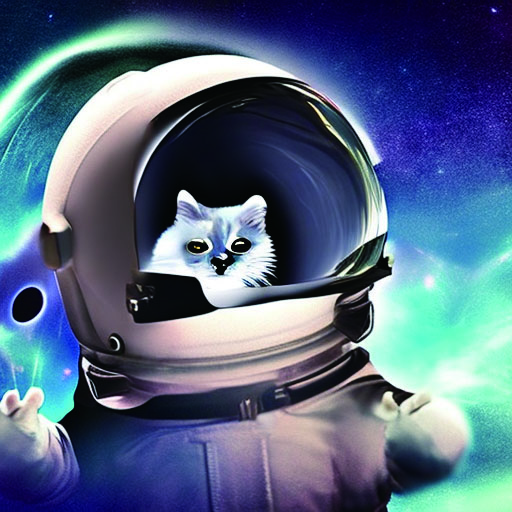} % HSO LAION result
    \caption{HSO+UniPC.}
    \label{fig:qual_HSO_unipc_laion}
\end{subfigure}
\hfill
\begin{subfigure}{0.3\linewidth}
    \includegraphics[width=\linewidth]{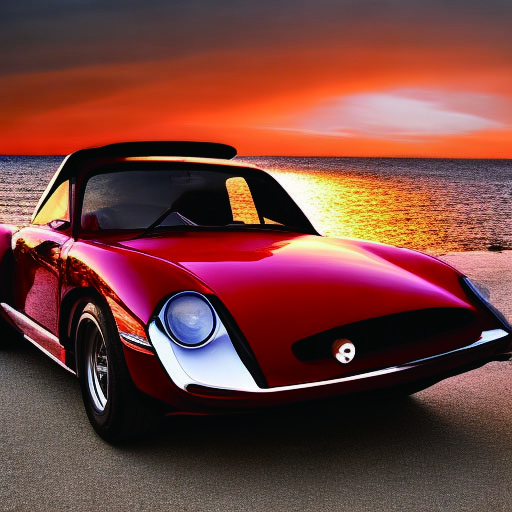} % HSO COCO result
    \caption{HSO+DDIM.}
    \label{fig:qual_HSO_ddim_coco}
\end{subfigure}
\hfill
\begin{subfigure}{0.3\linewidth}
    \includegraphics[width=\linewidth]{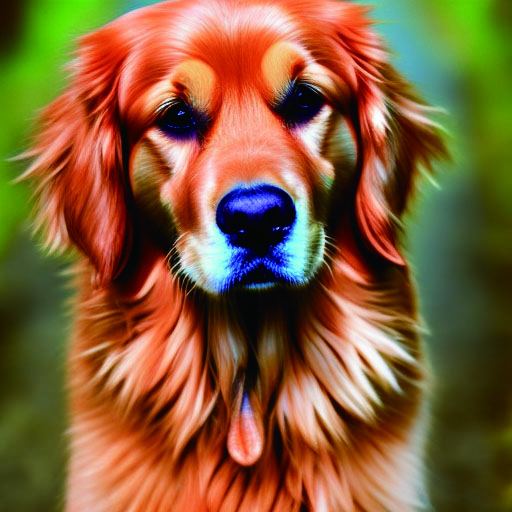} % HSO ImageNet result
    \caption{HSO+UniPC.}
    \label{fig:qual_HSO_unipc_imagenet}
\end{subfigure}

\begin{subfigure}{0.31\linewidth}
    \includegraphics[width=\linewidth]{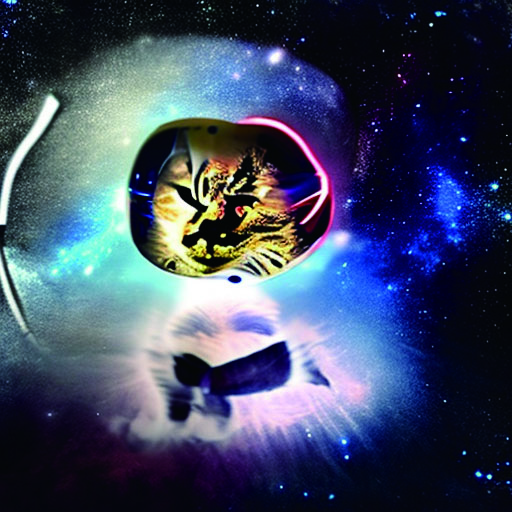} 
    \caption{baseline+UniPC.}
    \label{fig:qual_base_unipc_laion}
\end{subfigure}
\hfill
\begin{subfigure}{0.31\linewidth}
    \includegraphics[width=\linewidth]{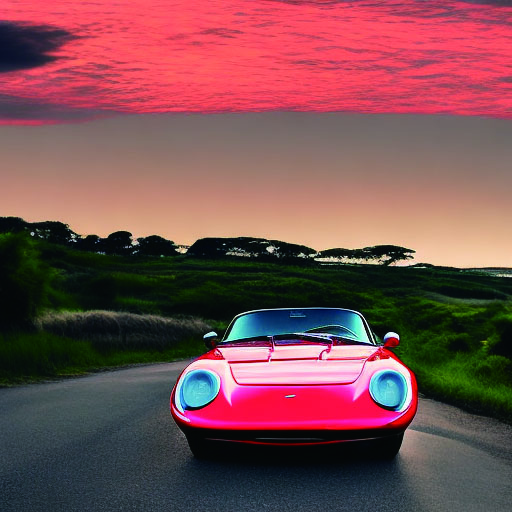} 
    \caption{baseline+DDIM.}
    \label{fig:qual_base_ddim_coco}
\end{subfigure}
\hfill
\begin{subfigure}{0.31\linewidth}
    \includegraphics[width=\linewidth]{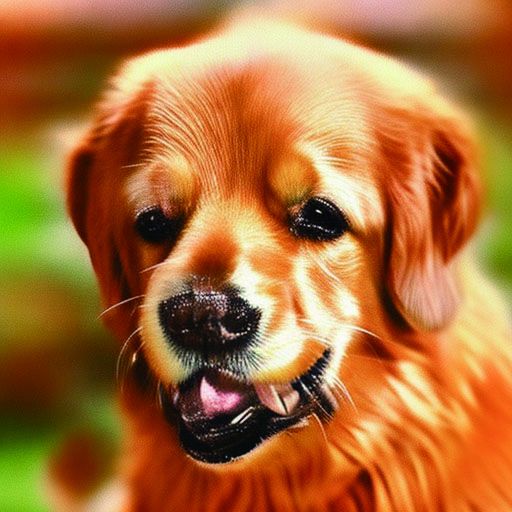} 
    \caption{baseline+UniPC.}
    \label{fig:qual_base_unipc_imagenet}
\end{subfigure}
\caption{Visual comparison of HSO (top) vs. DM-NonUni (bottom). HSO demonstrates superior image quality and coherence, especially at extremely low NFEs.}
\end{figure}

\textbf{Qualitative Analysis.} 
The qualitative results in Figure 1 visually corroborate our quantitative advantages over baseline DM-NonUni. HSO consistently generates images with high semantic accuracy and detail fidelity across different datasets and solvers. For instance, it accurately renders the ``astronaut cat" (Fig. \ref{fig:qual_HSO_unipc_laion}), the ``vintage car on a coast" (Fig. \ref{fig:qual_HSO_ddim_coco}), and a realistic ``golden retriever" (Fig. \ref{fig:qual_HSO_unipc_imagenet}). In contrast, the baseline method struggles at low NFEs, leading to both severe artifacts and semantic failures. This is evident in the distorted cat-helmet fusion (Fig. \ref{fig:qual_base_unipc_laion}), the failure to generate the correct coastal background for the car (Fig. \ref{fig:qual_base_ddim_coco}), and the malformed retriever face (Fig. \ref{fig:qual_base_unipc_imagenet}).
 
\subsection{Validation of Adaptivity}
\label{sec:adaptivity_validation}
HSO demonstrates strong adaptivity to both computational budgets (NFE) and model noise schedules.

\textbf{NFE Adaptivity.}
\label{sec:nfe_adaptivity}
We hypothesize the optimization landscape is highly non-convex, making any fixed rule-based approach unlikely to be optimal across all scenarios. The optimization-based nature for HSO maked it able to tailor the strategy adaptively according to the given. To validate this, we tasked HSO to find the optimal parameters $\boldsymbol{\psi}^*$ for various NFE values. The results, presented in Table \ref{table:nfe_adaptive}, strikingly demonstrate this adaptive behavior. The optimal parameters exhibit a complex, non-monotonic relationship with NFE (e.g., $\rho^*$ peaks at NFE=8), refuting any ``one-size-fits-all" approach and highlighting the necessity of HSO's adaptive search. 

% The correspondence between $\boldsymbol{t}$ and the $\boldsymbol{\sigma}$ domain is detailed in Appendix~C.

\begin{table}[htbp]
\centering
\small
\begin{tabular}{
    S[table-format=2.0]
    S[table-format=1.8]
    S[table-format=1.8]
    S[table-format=2.4]
    S[table-format=2.2]
}
\toprule
{\textbf{NFE}} & {$\boldsymbol{t_{\min}^*}$} & {$\boldsymbol{t_{\max}^*}$} & {$\boldsymbol{\rho^*}$} & {\textbf{FID}} \\
\midrule
4 & 0.03       & 0.96       & 8.8431 & 15.71 \\
6 & 0.01000187 & 0.9999368  & 6.5085 & 11.17 \\
8 & 0.01000035 & 0.9998941  & 12.4163& 8.79 \\
10& 0.0102511  & 0.9999527  & 11.6272& 8.32 \\
\bottomrule
\end{tabular}
\caption{Optimal initialization parameters discovered by HSO for varying NFEs. We search for $t$ instead of $\sigma$, as they share a one-to-one correspondence (Appendix C).}
\label{table:nfe_adaptive}
\end{table}

\textbf{Model Adaptivity.}
\label{sec:model_adaptivity}
HSO's adaptivity also extends to the unique noise characteristics inherent in different models. Theoretically, any two models with distinct noise progressions (i.e., different mappings from timestep t to noise levels $\alpha_t$ and $\sigma_t$) require different optimal sampling schedules, even if they belong to the same model family. To validate this, we test HSO on PixArt-$\alpha$ on the MS-COCO 30k validation set, using the UniPC solver with an NFE of 5. Although both PixArt-$\alpha$ and Stable Diffusion v2.1 are Variance-Preserving (VP) models, they employ different $\beta$ schedules, resulting in distinct model properties that HSO can adapt to. As shown in Table \ref{tab:model_adaptivity}, applying HSO to PixArt-$\alpha$ significantly improves the FID score over its default schedule. This result demonstrates that HSO effectively optimizes for the specific characteristics of a given model, underscoring its versatility.

\begin{table}[htbp]
\centering
\small
\renewcommand{\arraystretch}{1.2}
\begin{tabular*}{\columnwidth}{
  @{}
  l % Left-aligned column for model name
  @{\extracolsep{\fill}} % Adds flexible space between columns
  S[table-format=2.2]   % For baseline FID
  S[table-format=2.2] % For HSO FID
  @{}
}
\toprule
\textbf{Foundational Model} & {\textbf{Baseline FID}} & {\textbf{HSO FID}} \\
\midrule
PixArt-$\alpha$ (512px) & 37.65 & 18.05 \\
SD-V2.1 (512px) & 24.06 & 19.59 \\
\bottomrule
\end{tabular*}
\caption{HSO performance on PixArt-$\alpha$ (512px), evaluated with NFE=5 and the UniPC solver on the MS-COCO 30k validation set.}
\label{tab:model_adaptivity}
\end{table}

\subsection{Validation of Practical Robustness}
\label{sec:robustness_validation}

In the following experiments with the DDIM scheduler, we use the standard discretization where the continuous time range is mapped to 1000 integer steps (indexed 0-999). Unconstrained optimization can yield ``brittle'' schedules with pathologically close timesteps that hurt empirical performance. For instance, without our SPF, the DDIM schedule collapsed to \texttt{[999, 70, 9, 9]}, exhibiting an extreme ``end-clustering'' of steps, resulting in a catastrophic FID of 165.48 (Table \ref{tab:robustness_comparison}). The minimum spacing between steps collapsed to 0.0, a clear indicator of numerical instability, which resulted in an unusable FID of \textbf{165.48}. In contrast, enabling our NFE-adaptive SPF penalty enforces a healthy minimum step spacing (243.0) and reliably guides the optimizer toward high-performing, stable solutions, achieving a stable mean FID of \textbf{19.76±0.25} over 10 runs. This confirms our regularization is crucial for practical robustness.

\begin{table}[htbp]
\centering
\small
\renewcommand{\arraystretch}{1.4}

\begin{tabular}{@{} >{\raggedright\arraybackslash}p{0.16\columnwidth} >{\raggedright\arraybackslash}p{0.24\columnwidth} >{\centering\arraybackslash}p{0.22\columnwidth} >{\centering\arraybackslash}p{0.21\columnwidth} @{}}
\toprule
\textbf{Condition} & \textbf{Example Schedule} & \textbf{Step Spacing (Min / Mean)} & \textbf{FID} \\
\midrule
Without Penalty & [999, 70, 9, 9] & 0.0 / 330.0 & \textbf{165.48} \newline (Unstable) \\
\midrule
With Penalty & [959, 716, 370, 30] & 243.0 / 309.7 & \textbf{19.76 $\pm$ 0.25} \newline (\(\sim\)10 runs) \\
\bottomrule
\end{tabular}
\caption{Comparison of schedules optimized with and without a penalty. The penalty-based regularization prevents step collapse and leads to stable, high-quality results.}
\label{tab:robustness_comparison}
\end{table}

\subsection{Validation of the Computational Efficiency}
\label{sec:computational_efficiency}

A core design principle of HSO is practical computational efficiency. This section validates HSO's \textbf{preparation cost} against other leading ``Training-Free" and ``Training-Based" methods, as shown in Table~\ref{tab:sota_comparison}.
HSO's efficiency is a key advantage, finding its schedule in \(\sim\)8 seconds on a consumer-grade CPU. Within the training-free landscape, while slightly longer than the purely local search of DM-NonUni (\(\sim\)1s), this is an expected trade-off for our bi-level global search and is orders of magnitude faster than other global methods like AutoDiffusion (\(\sim\)1.1 days). Crucially, this low cost is paired with a leading FID score on the challenging ImageNet 512x512 benchmark, whereas competitors often report on lower-resolution 256x256 images. 
As a training-free method, HSO's 8-second cost bypasses the substantial training investments (often GPU-days or months) required by training-based paradigms.
% It is also worth noting that, by operating in the training-free paradigm, HSO bypasses the substantial training investments (often GPU-days or months) required by training-based methods. This makes high-fidelity, few-step generation practical and accessible without large-scale training resources.

\begin{table}[htbp]
\centering
\small
\renewcommand{\arraystretch}{1.4}
\setlength{\tabcolsep}{0pt}
\newcolumntype{M}[1]{>{\centering\arraybackslash}m{#1}}

\begin{tabularx}{\columnwidth}{
  @{}
  M{1.5cm} % Col 1 (Group)
  @{\hspace{0.2pt}} % Gap 1-2
  >{\raggedright}p{2.1cm} % Col 2 (Method)
  @{\hspace{0.2pt}} % Gap 2-3
  >{\raggedright\arraybackslash}X % Col 3 (Paradigm)
  @{\hspace{0.1pt}} % Gap 3-4
  M{1.1cm} % Col 4 (Prep. Cost)
  @{\hspace{1pt}} % Gap 4-5
  M{0.6cm} % Col 5 (NFE) - NEW
  @{\hspace{1pt}} % Gap 5-6
  S[table-format=2.2, table-space-text-post=\textsuperscript{\textsection}] % Col 6 (FID)
  @{}
}
\toprule
& \textbf{Method} & \textbf{Paradigm} & \textbf{Prep. Cost}\textsuperscript{$\ddagger$}& \textbf{NFE} & {\textbf{FID}} \\
\midrule
% --- Training-Free Methods ---
\multirow{3}{=}[-12pt]{\itshape\bfseries Training-Free Methods} 
& \textbf{HSO (Ours)} & \textbf{Hierarchical Opt.} & \textbf{\(\sim\)8s} & \textbf{4} & \bfseries 15.71\textsuperscript{\textdagger} \\
& AutoDiffusion \\ {[ICCV '23]} & Evolutionary Search & \(\sim\)1.1 d & 4 & 17.86 \\
& DM-NonUni \\ {[CVPR '24]} & Local Optimization & \(\sim\)1s & 4 & 18.96 \\
\midrule
% --- Training-Based Methods ---
\multirow{3}{=}[-15pt]{\itshape\bfseries Training-Based Methods} 
& UFOGen \\ {[CVPR '24]} & Diffusion-GAN & \(\sim\)12 d & 1 & 22.5\textsuperscript{*} \\
& Mean Flows \\ {[ICLR '25]} & Flow Matching & \(\sim\)60 d & 1 & 3.43\textsuperscript{**} \\
& LCM \\ {[ICCV '23]} & Distillation & \(\sim\)1.33 d & 4 & 11.10\textsuperscript{\textsection} \\
\bottomrule
\end{tabularx}

\parbox{\columnwidth}{\footnotesize
\vspace{0.5ex}
\raggedright
\textsuperscript{\mbox{$\ddagger$}}\textbf{
% s vs. d for magnitude comparison only.
Units are seconds (s) and days (d) to emphasize the magnitude of difference in cost.
} 
FID on ImageNet 256$\times$256 unless noted; 
\textsuperscript{\textdagger}ImageNet 512$\times$512; 
\textsuperscript{*}MSCOCO; 
\textsuperscript{**}ImageNet 256$\times$256;
\textsuperscript{\textsection}LAION-Aesthetics (SD v2.1).
}

\caption{Performance comparison with SOTA methods. Prep. cost is estimated in A100 GPU-days (d), except for HSO and DM-NonUni, which were measured on an Intel Core i7-14650HX.}
\label{tab:sota_comparison}
\end{table}

\subsection{Ablation Study}
We conducted an ablation study on the LAION-Aesthetics 6.5+ dataset with NFE=4 to isolate the contributions of HSO's key components: the Bi-Level search and the MEP objective. 
All experiments here include the SPF penalty, with its ablation shown in Section \ref{sec:robustness_validation}.
Results are detailed in Table \ref{tab:ablation_study_final_v3}.

\textbf{The baseline (A):} The principled local schedule optimization approach proposed by \citeauthor{xue2024accelerating}(\citeyear{xue2024accelerating}), which is an objective function designed specifically for solvers like Unipc.

\textbf{A+Bi-level (B):} Adding only our Bi-Level global search finds a superior initial schedule. This dramatically improves FID for both solvers, especially for DDIM ($71.57 \rightarrow 29.22$), as its default performance is far from optimal, offering more room for improvement from a better initialization compared to the already well-performing UniPC.

\textbf{A + MEP Objective (C):} This configuration replaces the baseline's UniPC-specific local objective with our solver-agnostic MEP. This reveals a predictable trade-off: the acceptable performance decrease on UniPC ($18.07 \rightarrow 26.33$) is outweighed by a massive improvement on DDIM ($71.57 \rightarrow 37.27$). This result strongly validates MEP's effectiveness as a universal error proxy that avoids overfitting to a single solver.

\textbf{Synergy in the Full HSO Framework (Ours):} The complete HSO framework demonstrates powerful synergy. The Bi-Level stage provides a superior foundation, which the MEP objective then refines to achieve the best overall performance. While the UniPC FID is slightly higher than in configuration (B) due to the objective trade-off described in (C), the final DDIM FID reaches a new best of 24.80. This proves our hierarchical design is essential for discovering robust, high-performing schedules.

\begin{table}[H]
    \centering
    \small 
    \renewcommand{\arraystretch}{1.3}
    
    \begin{tabular*}{\columnwidth}{
      @{}
      l % Left-aligned column for the text
      @{\extracolsep{\fill}} % Adds flexible space between columns
      S[table-format=2.2]
      % ####################################################################
      % ###                                                              ###
      % ###   V V V V V   THIS IS THE LINE YOU MUST CHANGE   V V V V V   ###
      % ###                                                              ###
      S[table-format=2.2, detect-weight] 
      % ####################################################################
      @{}
    }
        \toprule

        \textbf{Component / Configuration} & {\textbf{UniPC FID}} & {\textbf{DDIM FID}} \\
        \midrule
        
        Baseline (A)  & 18.07 & 71.57 \\
        \midrule
        
        A + Bi-Level (B) & 11.44 & 29.22 \\
        A + MEP Objective (C) & 26.33 & 37.27 \\
        \midrule

        % And here, use \bfseries to make the number bold
        Synergistic Framework (Ours HSO) & 15.70 & {\bfseries 24.80} \\ 
        \bottomrule
    \end{tabular*}
    \caption{Ablation study of HSO's components on UniPC and DDIM solvers (NFE=4).}
    \label{tab:ablation_study_final_v3}
\end{table}

\section{Conclusion}
\label{sec:conclusion}
In this paper, we addressed the challenge of accelerating diffusion models via training-free schedule optimization. We proposed the Hierarchical-Schedule-Optimizer (HSO), a novel bi-level framework designed to holistically satisfy the principles of Adaptivity, Effectiveness, Practical Robustness, and Computational Efficiency. HSO navigates the non-convex search landscape by synergizing a low-dimensional global search with a rapid local refinement, a process powered by our two innovations: the solver-agnostic Midpoint Error Proxy (MEP) objective and the stability-enhancing Spacing-Penalized Fitness (SPF) function. Extensive experiments validate that HSO establishes a new state-of-the-art in the low-NFE regime with a negligible optimization cost of just a few seconds, confirming it as a practical and powerful solution. For future work, the HSO framework could be extended to co-optimize other sampling components, such as solver hyperparameters, in a unified manner.

\section*{Acknowledgments}
This work is funded by the Science and Technology Development Fund, 
Macau SAR (File No. 0008/2025/RIB1, 0126/2025/RIA2, and 0077/2025/RIA2).

\bibliography{aaai2026}

\newpage
\appendix
\setcounter{lemma}{0}
\setcounter{theorem}{0}
\section{Proof of Lemma 1}

\begin{lemma}[Hybrid Midpoint Approximation]
The integral term can be approximated as: 
\begin{equation}
\label{eq:integral_approx}
    \int_{\lambda_i}^{\lambda_{i+1}}e^\lambda f(\lambda)d\lambda \approx f(\lambda_{i+\frac{1}{2}})(e^{\lambda_{i+1}}-e^{\lambda_i})
\end{equation}
where $\lambda_{i+\frac{1}{2}}\overset{\triangle}{=} \frac{\lambda_i+\lambda_{i+1}}{2}$. The local truncation error of this approximation is $O(h^3)$, the same order as the standard midpoint rule, but it is numerically more stable for this specific integral structure.
\label{lemma:approx}
\end{lemma}

\begin{proof}
For providing the complete proof for Lemma \ref{lemma:approx}, We first detail the derivation of the Hybrid Midpoint Approximation formula. Subsequently, we analyze its local truncation error via Taylor expansion to prove its third-order accuracy. Finally, we compare it with the standard midpoint rule to highlight our method's advantage in numerical stability while maintaining the same order of accuracy.

\subsection{Derivation of the Approximation}
The core principle of our Hybrid Midpoint Approximation is to separate the integral into two distinct components: an analytically integrable exponential term, $e^\lambda$, and the neural network output term, $f(\lambda)$. We treat $f(\lambda)$ as a smooth function that varies slowly over a small integration interval. This assumption is reasonable because, while individual components like activation functions (e.g., ReLU) can be non-smooth, the macroscopic output of a deep neural network tends to exhibit smooth behavior. Furthermore, $f(\lambda)$ represents the model's prediction of the clean data $x_0$ along the denoising trajectory, which is expected to evolve smoothly. Based on this, instead of approximating the entire integrand $e^\lambda f(\lambda)$ as the standard midpoint rule does, we only approximate $f(\lambda)$ as being constant over the small interval $[\lambda_i, \lambda_{i+1}]$. We evaluate it at the midpoint $\lambda_{i+\frac{1}{2}}$ and move it outside the integral. The remaining exponential term is then integrated exactly.

This leads to the following derivation:
\begin{equation}
\begin{aligned}
    \int_{\lambda_i}^{\lambda_{i+1}}e^\lambda f(\lambda)d\lambda &\approx f(\lambda_{i+\frac{1}{2}})\int_{\lambda_i}^{\lambda_{i+1}}e^\lambda d\lambda \\
    &= f(\lambda_{i+\frac{1}{2}})[e^\lambda]_{\lambda_i}^{\lambda_{i+1}}\\
    &= f(\lambda_{i+\frac{1}{2}})(e^{\lambda_{i+1}}-e^{\lambda_i})
\end{aligned}
\end{equation}

\subsection{Local Truncation Error Analysis}
To determine the order of the error, we analyze the local truncation error, $E_i$, which is the difference between the exact integral and our approximation over the interval $[\lambda_i, \lambda_{i+1}]$. The error of this approximation is the integral of the difference:
\begin{equation}
    E_i = \int_{\lambda_i}^{\lambda_{i+1}} e^{\lambda} [f(\lambda) - f(\lambda_{i+\frac{1}{2}})] d\lambda
\end{equation}

To analyze this error term, we perform a two-step Taylor expansion. Let $h=\lambda_{i+1}-\lambda_i$. First, we expand $f(\lambda)$ around the midpoint $\lambda_{i+\frac{1}{2}}$:

\begin{equation}
\begin{aligned}
    f(\lambda) &= f(\lambda_{i+\frac{1}{2}}) + f'(\lambda_{i+\frac{1}{2}})(\lambda - \lambda_{i+\frac{1}{2}}) \\
    &+ \frac{f''(\lambda_{i+\frac{1}{2}})}{2}(\lambda - \lambda_{i+\frac{1}{2}})^2 + O((\lambda - \lambda_{i+\frac{1}{2}})^3)
\end{aligned}
\end{equation}
Then, we expand $e^\lambda$ around $\lambda_{i+\frac{1}{2}}$:
\begin{equation}
\begin{aligned}
    e^\lambda = e^{\lambda_{i+\frac{1}{2}}}[1+(\lambda - \lambda_{i+\frac{1}{2}})+O((\lambda - \lambda_{i+\frac{1}{2}})^2)]
\end{aligned}
\end{equation}
Substituting these into the error integral, we expand the product and focus on terms up to the third order of $(\lambda - \lambda_{i+\frac{1}{2}})$:
\begin{equation}
\begin{aligned}
    E_i = &\int_{\lambda_i}^{\lambda_{i+1}} e^{\lambda_{i+\frac{1}{2}}}[1+(\lambda - \lambda_{i+\frac{1}{2}})+O((\lambda - \lambda_{i+\frac{1}{2}})^2)]\\
    &(f'(\lambda_{i+\frac{1}{2}})(\lambda - \lambda_{i+\frac{1}{2}})+\frac{f''(\lambda_{i+\frac{1}{2}})}{2}(\lambda - \lambda_{i+\frac{1}{2}})^2 + \\
    &O((\lambda - \lambda_{i+\frac{1}{2}})^3)) d\lambda\\
    =&e^{\lambda_{i+\frac{1}{2}}}\int_{\lambda_i}^{\lambda_{i+1}}[f'(\lambda_{i+\frac{1}{2}})(\lambda - \lambda_{i+\frac{1}{2}}) + \\
    &(f'(\lambda_{i+\frac{1}{2}})+\frac{f''(\lambda_{i+\frac{1}{2}})}{2})(\lambda - \lambda_{i+\frac{1}{2}})^2 + \\
    & O((\lambda - \lambda_{i+\frac{1}{2}})^3)]d\lambda
\end{aligned}
\end{equation}

We now integrate the expanded expression. Since the integration is over a symmetric interval $[\lambda_{i+\frac{1}{2}}-h/2, \lambda_{i+\frac{1}{2}}+h/2]$, any term with an odd power of $(\lambda-\lambda_{i+\frac{1}{2}})$ will integrate to zero. This means the terms of first order $(\lambda-\lambda_{i+\frac{1}{2}})$ and third order $O((\lambda-\lambda_{i+\frac{1}{2}})^3)$ vanish. The lowest-order non-zero term is therefore the second-order term. The next significant contributions come from the fourth-order term in the integrand, which could be considered as an integration of a generic fourth-order term, where C is a constant:
\begin{equation}
\begin{aligned}
    \int_{\lambda_i}^{\lambda_{i+1}} C (\lambda-\lambda_{i+\frac{1}{2}})^4 d\lambda &= C[\frac{(\lambda-\lambda_{i+1/2})^5}{5}]_{\lambda_i}^{\lambda_{i+1}}\\
    &=\frac{C}{5}[(\frac{h}{2})^5-(-\frac{h}{2})^5]\\
    &=\frac{C}{80}h^5=O(h^5)
\end{aligned}
\end{equation}
This demonstrates that the subsequent remainder terms are of order $O(h^5)$ or higher, allowing us to safely truncate the analysis at the second-order term to find the leading error.

Thus, the leading term of the error integral is:

\begin{equation}
\begin{aligned}
 E_i =&\int_{\lambda_i}^{\lambda_{i+1}}[e^{\lambda_{i+\frac{1}{2}}}(f'(\lambda_{i+\frac{1}{2}})+\frac{f''(\lambda_{i+\frac{1}{2}})}{2})(\lambda - \lambda_{i+\frac{1}{2}})^2]d\lambda\\
 +& O(h^5)\\
 =&e^{\lambda_{i+\frac{1}{2}}}(f'(\lambda_{i+\frac{1}{2}})+\frac{f''(\lambda_{i+\frac{1}{2}})}{2})[\frac{(\lambda - \lambda_{i+\frac{1}{2}})^3}{3}]_{\lambda_i}^{\lambda_{i+1}}\\
 +& O(h^5)\\
 =&e^{\lambda_{i+\frac{1}{2}}}(f'(\lambda_{i+\frac{1}{2}})+\frac{f''(\lambda_{i+\frac{1}{2}})}{2})\frac{h^3}{12}+O(h^5)
\end{aligned}
\end{equation}
Since the leading error term is proportional to $h^3$, the local truncation error is of order $O(h^3)$.

\subsection{Comparison with Standard Midpoint Rule
}
The standard midpoint rule approximates the entire integrand $e^\lambda f(\lambda)$ at the midpoint:
\begin{equation}
    \int_{\lambda_i}^{\lambda_{i+1}} e^{\lambda}f(\lambda) d\lambda \approx h \cdot e^{\lambda_{i+\frac{1}{2}}} f(\lambda_{i+\frac{1}{2}})
\end{equation}
To analyze its error in a manner consistent with our previous analysis, we can perform a Taylor expansion of the integrand $e^\lambda f(\lambda)$ around the midpoint $\lambda_{i+\frac{1}{2}}$. The error of the standard midpoint rule, $E_i^{(std)}$, is the integral of the Taylor expansion minus the approximation itself. The first two terms of the expansion, the constant and the linear term, integrate to the approximation value and zero, respectively. The leading error term therefore comes from integrating the second-order term of the expansion:

\begin{equation}
\begin{aligned}
    E_i^{(std)} = &\frac{h^3}{24}e^{\lambda_{i+\frac{1}{2}}}(f(\lambda_{i+\frac{1}{2}})+2f'(\lambda_{i+\frac{1}{2}})+f''(\lambda_{i+\frac{1}{2}}))\\
    &+ O(h^5)
\end{aligned}
\end{equation}

This confirms that the error is of order $O(h^3)$. While both methods have the same error order $O(h^3)$, our hybrid approach offers superior numerical robustness, which can be seen by comparing the leading error coefficients. The coefficient for our method's error, $E_i$, is $\frac{e^{\lambda_{i+\frac{1}{2}}}}{12}(f'(\lambda_{i+\frac{1}{2}})+\frac{f''(\lambda_{i+\frac{1}{2}})}{2})$, while the coefficient for the standard midpoint rule's error, $E_i^{(std)}$, is $\frac{e^{\lambda_{i+\frac{1}{2}}}}{24}(f(\lambda_{i+\frac{1}{2}})+2f'(\lambda_{i+\frac{1}{2}})+f''(\lambda_{i+\frac{1}{2}}))$. The key difference is that the error for the standard rule includes the term $e^{\lambda_{i+\frac{1}{2}}}f(\lambda_{i+\frac{1}{2}})$. As the sampling process approaches the clean image, the log-SNR $\lambda$ increases, causing the exponential term $e^{\lambda_{i+\frac{1}{2}}}$ to grow significantly. The presence of this large, location-dependent exponential term multiplied by the function value $f(\lambda_{i+\frac{1}{2}})$ itself makes the error coefficient of the standard rule potentially large and unstable. In contrast, our method's error coefficient only depends on the derivatives of $f(\lambda)$. By integrating the exponentially growing term $e^\lambda$ exactly, our method's error is isolated from the large magnitude of the integrand, resulting in a smaller and more stable error constant in practice.

\end{proof}

\section{Proof of Theorem 1}

\begin{theorem}[Global Error Bound for Hybrid Midpoint Approximation]
\label{theo:error}
Under the aforementioned assumptions, let the approximate solution $\tilde{x}_{\epsilon,MEP}$ be derived using the approximation from Lemma  \ref{lemma:approx}. The global generation error is then bounded with high probability by:
    \begin{equation}
        ||\tilde{x}_{\epsilon,MEP}-x_0||\leq C + \sigma_\epsilon\tilde{\eta}\sum_{i=0}^{N-1}\tilde{\epsilon}_{t(\lambda_{i+\frac{1}{2}})}(e^{\lambda_{i+1}}-e^{\lambda_i})
    \end{equation}
    where $C$ represents schedule-independent error terms.
\end{theorem}

\begin{proof}
We begin by defining the approximate final sample $\tilde{x}_{\epsilon,MEP}$ using our Hybrid Midpoint Approximation from Lemma \ref{lemma:approx}. To formalize this bound, we adopt the theoretical setup from DM-NonUni \citep{xue2024accelerating}, assuming the model's prediction is $f(\lambda)=x_0+\xi_{t(\lambda)}$, where the error norm is bounded with high probability by $||\xi_{t(\lambda)}||\leq\tilde{\eta}\tilde{\epsilon}_{t(\lambda)}$ for some constant $\tilde{\eta}$ and error proxy $\tilde{\epsilon}_{t(\lambda)}=\frac{\sigma_t^p}{\alpha_t}$ for a non-negative integer $p$. The following derivation bounds the norm of the global error, $||\tilde{x}_{\epsilon,MEP}-x_0||$.

The derivation proceeds by substituting the definition of $\tilde{x}_{\epsilon,MEP}$, applying the model prediction assumption $f(\lambda)=x_0+\xi_{t(\lambda)}$, and using the triangle inequality to separate the error terms. The final bound consists of two parts: a schedule-independent error term C, which depends only on the start and end points of the integration, and a schedule-dependent summation.

\begin{equation}
\begin{aligned}
   ||\tilde{x}_{\epsilon,MEP}&-x_0||\\
   &=||\frac{\sigma_\epsilon}{\sigma_T}x_T+\sigma_\epsilon\sum_{i=0}^{N-1}f(\lambda_{i+\frac{1}{2}})(e^{\lambda_{i+1}}-e^{\lambda_i})-x_0|| \\
   &=||\frac{\sigma_\epsilon}{\sigma_T}x_T+\sigma_\epsilon\sum_{i=0}^{N-1}(x_0 +\xi_{i+\frac{1}{2}})(e^{\lambda_{i+1}}-e^{\lambda_i})-x_0||\\
   &=||\frac{\sigma_\epsilon}{\sigma_T}x_T+\sigma_\epsilon\sum_{i=0}^{N-1}x_0(e^{\lambda_{i+1}}-e^{\lambda_i})\\
   &\quad\quad +\sigma_\epsilon\sum_{i=0}^{N-1}\xi_{i+\frac{1}{2}}(e^{\lambda_{i+1}}-e^{\lambda_i})-x_0||\\
   &\leq ||\frac{\sigma_\epsilon}{\sigma_T}x_T+\sigma_\epsilon\sum_{i=0}^{N-1}x_0(e^{\lambda_{i+1}}-e^{\lambda_i})-x_0|| \\
   &\quad\quad+ \sigma_\epsilon\sum_{i=0}^{N-1}(e^{\lambda_{i+1}}-e^{\lambda_i})||\xi_{t(\lambda+\frac{1}{2})}||\\
   &\leq C + \sigma_\epsilon\tilde{\eta}\sum_{i=0}^{N-1}\tilde{\epsilon}_{t(\lambda_{i+\frac{1}{2}})}(e^{\lambda_{i+1}}-e^{\lambda_i})
\end{aligned}
\end{equation}
To minimize the overall error bound, we must minimize this summation. Therefore, the schedule-dependent part forms the basis of our MEP objective function, $\mathcal{J}_\text{MEP}$.
\end{proof}

\section{Hyperparameter Search Space and Notation Clarification}
Our framework optimizes an initial schedule generated by the formula proposed in EDM\cite{karras2022elucidating}. As described in Section 3.2 of our main paper, this schedule is governed by three key parameters: the distribution controller $\rho$, a minimum noise-level parameter $\tilde{\sigma}_{\min}$, and a maximum noise-level parameter $\tilde{\sigma}_{\max}$.

In practice, searching for the optimal noise-level parameters $(\tilde{\sigma}_{\min}, \tilde{\sigma}_{\max})$ is equivalent to searching for their corresponding time-domain values $(t_{\epsilon}, t_{\max})$. This is because the function $\tilde{\sigma}_t = \sigma_t/\alpha_t$ provides a monotonic mapping between the time $t$ and the noise-level parameter $\tilde{\sigma}_t$. For practical implementation, our upper-level search was therefore designed to optimize these time-domain values. Below, we describe the search space for the three parameters that are directly optimized by HSO's upper-level search.

The first parameter, $\rho$, controls the distribution of timesteps across the noise schedule. As explained by \citeauthor{karras2022elucidating}, a higher $\rho$ value concentrates more sampling steps at lower noise levels (i.e., higher SNR), which is perceptually more significant. While they note that $\rho=3$ provides a numerically-balanced schedule, values in the range of 5 to 10 often yield better results for image generation. Our chosen \textbf{search range of $\rho \in [3, 16]$} covers these findings and extends them, allowing our optimizer to explore even more aggressive schedules that heavily prioritize the final, high-detail stages of generation. This exploration is particularly relevant for the extremely low NFE scenarios that HSO is designed to address.

The other two parameters are the start time $t_{\max}$ and end time $t_{\epsilon}$ of the sampling process, which in turn determine the maximum and minimum noise-level parameters, $\tilde{\sigma}_{\max}$ and $\tilde{\sigma}_{\min}$. For the end time $t_\epsilon$, we set the \textbf{search range to $[0.01,0.03]$}. In the context of few-step generation, ending the process at an extremely small time t (e.g., the model's absolute minimum of 1/1000) is likely inefficient, as these final steps can consume a significant portion of the limited budget for imperceptible changes. We therefore hypothesize that a slightly larger end time might provide a better trade-off, and our search range allows the optimizer to test this. For the start time $t_{\max}$, we use a tight \textbf{search range of $[0.96,1.0]$}. A start time close to the maximum time $T=1.0$ ensures the process begins from a state of high entropy. This range was found empirically to provide a robust starting point for HSO, ensuring a high-entropy start without being excessively far from the data manifold.

\end{document}